\pdfoutput=1  

\documentclass{article}

\usepackage{arxiv}

\usepackage[T1]{fontenc}
\usepackage{amsmath}
\usepackage{amssymb}
\usepackage{graphicx}
\usepackage{hyperref}
\hypersetup{
    colorlinks=true,
    linkcolor=blue,
    citecolor=blue,
    urlcolor=blue,
}

\renewcommand{\headeright}{Accepted at ICRA 2026}
\renewcommand{\undertitle}{Accepted for publication at IEEE ICRA 2026}
\renewcommand{\shorttitle}{iVISION-2DCD: Long-Term Construction Change Detection}

\title{iVISION-2DCD: A Long-Term Change Detection Dataset for Large-Scale Outdoor Construction Monitoring
\thanks{This is the authors' version of a paper accepted for publication at the 2026 IEEE International Conference on Robotics and Automation (ICRA). The final version of record will be available via IEEE Xplore.}
}

\author{
    Dayou Mao\textsuperscript{1}\thanks{Corresponding authors: Dayou Mao (\texttt{daniel.mao@nrc-cnrc.gc.ca}), Ashkan Ebadi (\texttt{ashkan.ebadi@nrc-cnrc.gc.ca}), Yuhao Chen (\texttt{yuhao.chen1@uwaterloo.ca}).} \quad
    Yuchen Lin\textsuperscript{2} \quad
    Ashkan Ebadi\textsuperscript{1} \quad
    John Zelek\textsuperscript{2} \quad
    Alexander Wong\textsuperscript{2} \quad
    Yuhao Chen\textsuperscript{2} \\[2mm]
    \textsuperscript{1}National Research Council Canada, Ottawa, ON, Canada \\[1mm]
    \textsuperscript{2}Vision and Image Processing Research Group, Systems Design Engineering, \\[0.5mm]
    University of Waterloo, Waterloo, ON, Canada
}

\date{}

\begin{document}

\maketitle

\begin{abstract}

Automation in construction is essential for reducing costs and human errors in large-scale projects.
We approach the construction progress monitoring from the aspect of detecting changes in construction sites. As construction buildings continue to evolve in geometry and appearance over time, change detection need to be performed from \textit{arbitrary} camera viewpoints.
This necessitates developing 2D Change Detection (2DCD) algorithms that operate robustly across diverse camera perspectives at construction sites.
While developing and evaluating such systems is data-intensive, no open-source benchmark dataset exists at the intersection of 2D change detection and construction automation research.
Data collection using Unmanned Aerial Vehicles (UAVs) is gaining its popularity in outdoor large-scale surveying. However, in active construction sites conducting drone missions equipped with high-end sensors imposes safety concerns. Flight trajectory and collected camera viewpoints can be significantly limited.
To address this critical gap, we introduce iVISION-2DCD, a large-scale synthetically generated dataset from dense LiDAR point clouds with photorealistic input images and accurate ground truth annotations. Our dataset formally defines the problem of viewpoint-robust 2DCD at construction sites and captures the inherent complexities of real-world deployment.
In this paper, we present our systematic methodology for synthetic data generation, developing novel view synthesis techniques to overcome bi-temporal alignment and viewpoint diversity challenges, and implementing semi-automated semantic segmentation with change label generation while preserving challenging real-world cases.
Benchmark evaluations using state-of-the-art 2DCD algorithms demonstrate that iVISION-2DCD poses novel research challenges for the computer vision and robotics communities.
Project page: \url{https://danielmao2019.github.io/iVISION-2DCD-dataset.github.io/}.

\end{abstract}

\begin{figure}[t]
    \includegraphics[width=0.24\textwidth]{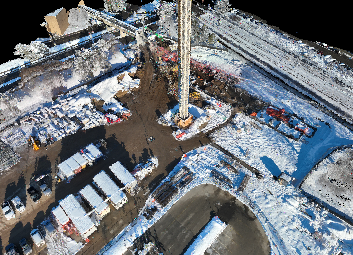}
    \includegraphics[width=0.24\textwidth]{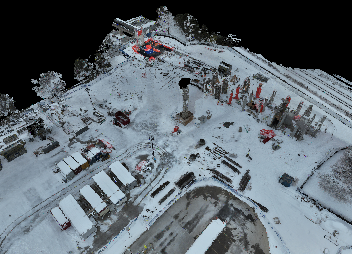}
    \includegraphics[width=0.24\textwidth]{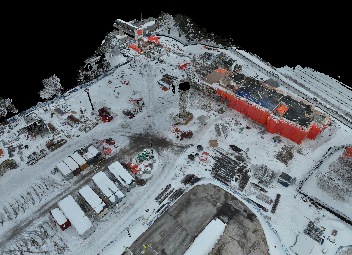}
    \includegraphics[width=0.24\textwidth]{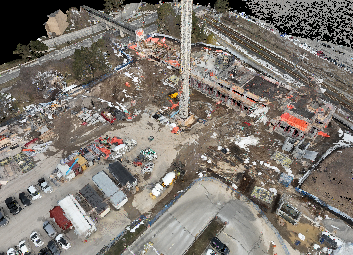}
    \includegraphics[width=0.24\textwidth]{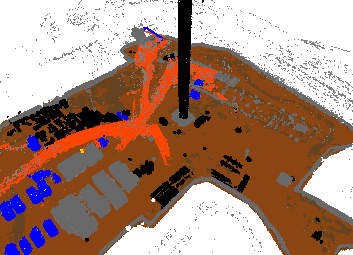}
    \includegraphics[width=0.24\textwidth]{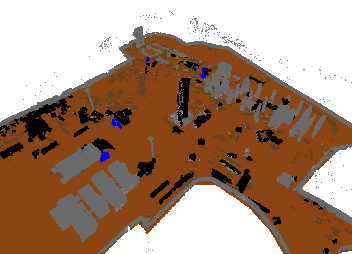}
    \includegraphics[width=0.24\textwidth]{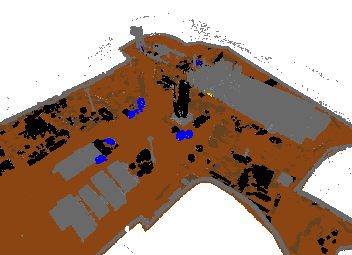}
    \includegraphics[width=0.24\textwidth]{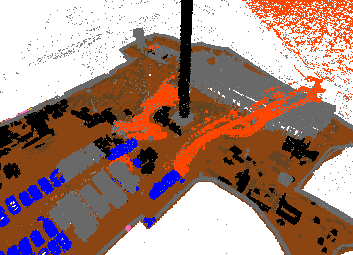}
    \includegraphics[width=0.24\textwidth]{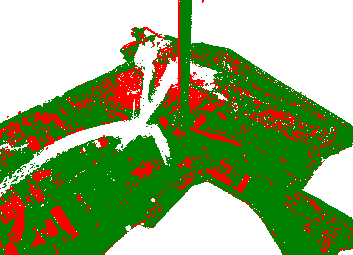}
    \includegraphics[width=0.24\textwidth]{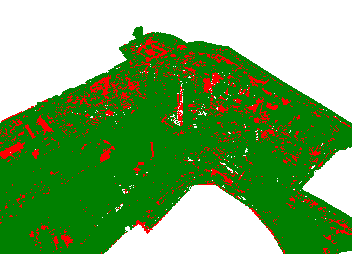}
    \includegraphics[width=0.24\textwidth]{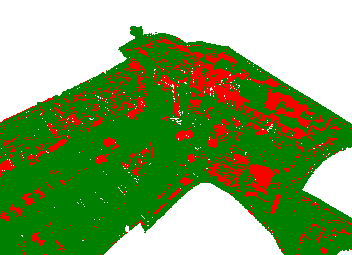}
    \includegraphics[width=0.24\textwidth]{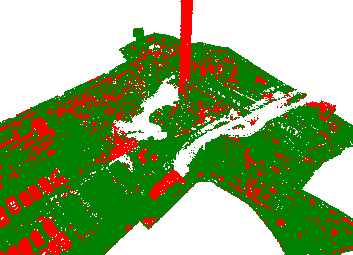}
    \caption{Example scenes (not consecutive) rendered from a distant camera: Dec-03-2024, Jan-19-2025, Feb-08-2025, Mar-12-2025, from left to right. First row: rendered RGB image from the LiDAR point cloud. Second row: rendered 3D segmentation point cloud from manual annotation. Third row: rendered 3D change point cloud compared from immediate previous scene. In the change maps, green represents unchanged points; red represents changed points; white represents ignored points.}
    \label{fig:first_figure}
\end{figure}

\section{INTRODUCTION}

The construction industry, representing approximately USD 10 trillion annually and comprising 13\% of global GDP, faces severe inefficiencies where over 53\% of projects experience delays and 66\% exceed budgets due to inadequate monitoring approaches~\cite{ham2016visual, agarwal2016imagining}. Manual monitoring methods contribute to schedule overruns of up to 20\% and cost escalations reaching 80\% above estimates~\cite{golparvar2015automated, yang2015construction}, highlighting the urgent need for automated solutions.
Change Detection (CD) algorithms enable automated progress monitoring by providing objective, quantitative measurements that circumvent time-consuming manual inspections and facilitate continuous monitoring across large spatial extents impractical for traditional survey methods~\cite{meyer2022change, suh2024monitoring, han2021change}.
Unmanned Aerial Vehicles (UAVs) provide efficient large-scale data acquisition through cost-effective monitoring~\cite{han2021change}, enhanced safety by eliminating ground-based personnel requirements~\cite{ham2016visual}, and affordable 3D reconstruction with rich semantic information~\cite{huang2022semantics}.
While 3D Change Detection (3DCD) offers geometric advantages, 2D Change Detection (2DCD) remains the practical choice for widespread deployment: RGB-only drones are significantly more cost-effective and lightweight compared to LiDAR-equipped systems, reducing safety constraints and operational expertise requirements that limit accessibility for construction teams. In this paper, we focus on 2DCD for construction progress monitoring.

While deep learning approaches have shown promise in this domain, they require substantial training data. Despite this need, existing datasets exhibit critical limitations: temporal sampling inadequacies with studies spanning merely 2.5-4 months and 3-5 timestamps, spatial resolution constraints limiting detection precision, and environmental bias toward controlled indoor scenarios that fail to capture outdoor complexities including seasonal variations and weather conditions~\cite{han2021change, huang2022semantics}, and single camera pose from top-down view captured by satellite imageries \cite{oscd,whu_cd,dsamnet}. There exists no comprehensive dataset for 2DCD in long-term, large-scale outdoor construction environments that addresses these fundamental limitations.

To address this critical gap in 2DCD datasets for construction monitoring, we introduce iVISION-2DCD and present a comprehensive methodology for curating this large-scale dataset.
Our data acquisition employs a DJI Matrice 350 drone equipped with RGB cameras and LiDAR sensors, enabling dense multi-modal capture.
Flight trajectories are planned to balance operational constraints and data quality. The dataset curation addresses the challenge of diverse viewpoint sampling through novel view synthesis from dense LiDAR point clouds, systematic camera generation, and pixel coverage optimization, creating bi-temporally aligned image pairs from arbitrary viewing angles.
We algorithmically generate change labels from semantic class annotations to overcome manual annotation costs, while preserving dataset authenticity by retaining challenging real-world cases: terrain changes invisible in RGB, dynamic object artifacts from construction equipment, and systematic projection errors.
Our methodology transforms raw multi-temporal acquisitions into a structured benchmark with diverse viewpoints and accurate change labels, capturing construction dynamics from subtle material additions to major structural transformations across varying environmental conditions.

Our resulting dataset distinguishes itself along the following six dimensions:
(i) high-fidelity data acquisition utilizing state-of-the-art UAV platforms with advanced imaging sensors;
(ii) synthetic 2DCD data generated from synthetically generated diverse camera angles;
(iii) long-term temporal coverage with over 50 acquisition sessions spanning one full year, capturing seasonal variations including winter snow conditions;
(iv) extensive spatial coverage spanning large construction sites (32.3K $m^{2}$);
(v) authentic complexity from active construction environments; and
(vi) multi-modal data incorporating both geometric and textural information.
We establish performance baselines for state-of-the-art 2DCD algorithms, demonstrating that iVISION-2DCD introduces novel challenges at the intersection of computer vision, robotics, and construction automation.

Our contribution in this paper is 3-fold:
\begin{itemize}
    \item We present a systematic methodology for generating synthetic 2DCD data from dense LiDAR point clouds, overcoming operational constraints through novel view synthesis and semi-automated annotation.
    \item We introduce iVISION-2DCD, a novel benchmark dataset that captures the unique complexities of real-world construction monitoring scenarios.
    \item We demonstrate that iVISION-2DCD establishes new research challenges on vision-based construction progress monitoring systems through comprehensive benchmark comparison of existing 2DCD algorithms and open-sourced datasets.
\end{itemize}
This work represents an ongoing research effort.
The work in this paper utilizes data from November 27, 2024 to March 12, 2025.
Raw drone data from September 2024 to August 2025 has been made public.

\section{RELATED WORK}

\subsection{Construction Datasets}

Datasets supporting research in automation in construction progress monitoring exhibit critical limitations across three dimensions: temporal sampling inadequacies that miss construction dynamics, environmental bias toward controlled indoor settings, and single modality dependencies that ignore complementary sensor information.

\textbf{Temporal Sampling Inadequacies.} Current outdoor construction datasets demonstrate critically sparse temporal sampling that fails to capture construction dynamics adequately. UAV-based road construction monitoring spans only 2.5 months with merely 3 temporal acquisitions~\cite{han2021change}, while UAV photogrammetric construction site studies cover 4 months with only 5 timestamps~\cite{huang2022semantics}. This temporal sparsity prevents detection of gradual construction changes and completely misses seasonal environmental variations essential for robust outdoor monitoring algorithms.


\textbf{Environmental Bias and Limited Outdoor Applicability.} A significant portion of construction change detection research demonstrates problematic bias toward indoor environments that fails to address outdoor complexities~\cite{meyer2022change, czerniawski2021automated}. Indoor construction monitoring faces constraints including limited views, weak scanning geometries, and occlusions while demanding higher geometric resolution~\cite{meyer2022change}, yet these controlled conditions cannot capture environmental challenges inherent in outdoor sites including seasonal vegetation variations, dynamic shadow patterns, and weather-induced visual changes.

\textbf{Single Modality Dependencies.} Many studies rely on single data modalities - either RGB images only, point clouds only, or SAR imagery only - missing complementary information available through multi-modal sensing approaches that combine geometric and radiometric information for robust environmental adaptation~\cite{han2021change, attard2018vision, yang2017monitoring}.

\subsection{Change Detection}
\label{subsec:change_detection}

We provide a brief literature review of Change Detection (CD) starting with the traditional bi-temporal framework,
then examining existing benchmark datasets and their limitations,
followed by two improved learning paradigms that address specific limitations:
single-temporal methods for reducing annotation costs
and 3D approaches for incorporating geometric information.

\textbf{Bi-Temporal CD.}
Typical formulation of (binary) 2DCD task is:
given bi-temporal, RGB images $I_{1}$ and $I_{2}$ captured by the same camera (with same camera intrinsics and pose),
we are to perform dense classification to predict class 0 to unchanged pixels, and class 1 to changed pixels \cite{fc_siam, snunet_cd, ftn, cd_mamba, change_mamba}.
Neural networks can be trained with supervision from ground-truth per-pixel change labels.
However, the initial research focus has two limitations -
high costs in dense bi-temporal change labels as compared to single-temporal labels;
and lack of depth information in 2D maps, which are respectively address by single-temporal CD and 3DCD research, as we describe in the following paragraphs.

\textbf{2DCD Datasets. }
Existing 2DCD benchmarks include satellite datasets (OSCD~\cite{oscd}, Sentinel-2, Landsat) with coarse resolution (10-30m), aerial imagery datasets (LEVIR-CD~\cite{levir_cd}, SYSU-CD~\cite{dsamnet}, WHU-CD) with higher resolution (0.5-2m) for urban monitoring, and disaster assessment datasets (xBD~\cite{xbd}, AirChange~\cite{airchange}).
These datasets share a critical limitation: exclusive reliance on nadir-view imagery from remote sensing platforms, leaving vertical structures unobserved, volumetric changes unquantified, and creating occlusion blind spots.
The absence of oblique or street-level perspectives severely limits their applicability to real-world scenarios requiring multi-angle change analysis.

\textbf{Single-Temporal CD. }
Single-temporal CD addresses the prohibitive expense of pairwise labeling bi-temporal images by reformulating CD from requiring temporally-paired images with pixel-perfect alignment to learning from arbitrary unpaired images, where any two single-temporal images can form a training pair regardless of their temporal or spatial correspondence~\cite{change_star_v1, i3pe}.
The mainstream approach constructs pseudo bi-temporal pairs by artificially introducing changes into single-temporal images and leveraging existing semantic annotations to automatically generate change labels, transforming the expensive bi-temporal annotation problem into a more tractable single-temporal semantic labeling task~\cite{change_star_v1, change_star_v2, i3pe, mtcnet, changeclip}.

\textbf{3DCD. }
A sub-area of CD \cite{siam_kpconv, tran2018integrated, zhan2024pgn3dcd, lu2025ms} design neural networks on point cloud CD datasets \cite{urb3dcd, slpccd} due to the well-known problem of lack of elevation information from 2D views.
In urban building construction application, this problem is especially pronounced.
Change map generation in 3D space is necessary before projecting into 2D for curating our iVISION-2DCD dataset.



\section{Synthetic Data Generation}
\label{subsec:generation}

Obtaining 2DCD data for construction monitoring faces three fundamental challenges: UAV missions cannot achieve pixel-perfect bi-temporal alignment, operational constraints limit viewpoint diversity, and manual change annotation proves prohibitively expensive. We develop novel view synthesis techniques to solve the bi-temporal alignment and viewpoint diversity challenges (Section~\ref{subsec:novel_view_synthesis}), and implement semi-automated semantic segmentation with change label generation to overcome the manual annotation cost challenge (Section~\ref{subsec:seg_and_change}). This pipeline transforms dense LiDAR point clouds into co-registered photorealistic imagery with accurate ground truth change labels capturing construction dynamics from perspectives impossible during actual drone missions.

\subsection{Novel View Synthesis}
\label{subsec:novel_view_synthesis}

Operational UAV missions must continuously optimize trajectories around evolving construction geometry - structures rise, and site layouts transform between captures. This makes it impossible to repeat exact flight paths and camera poses across temporal acquisitions. Yet typical 2DCD algorithms require pixel-perfect bi-temporal alignment with identical camera poses, fundamentally necessitating synthetic rendering from our LiDAR reconstructions.
However, synthetic rendering introduces two additional challenges we must address: limited viewpoint diversity from safety restrictions in real captures (Section~\ref{subsubsec:synthetic_camera_generation}) and pixel coverage gaps inherent in point cloud rendering (Section~\ref{subsubsec:pixel_coverage}).

\subsubsection{Synthetic Camera Generation}
\label{subsubsec:synthetic_camera_generation}

\begin{figure}
    \centering
    \includegraphics[width=0.24\linewidth]{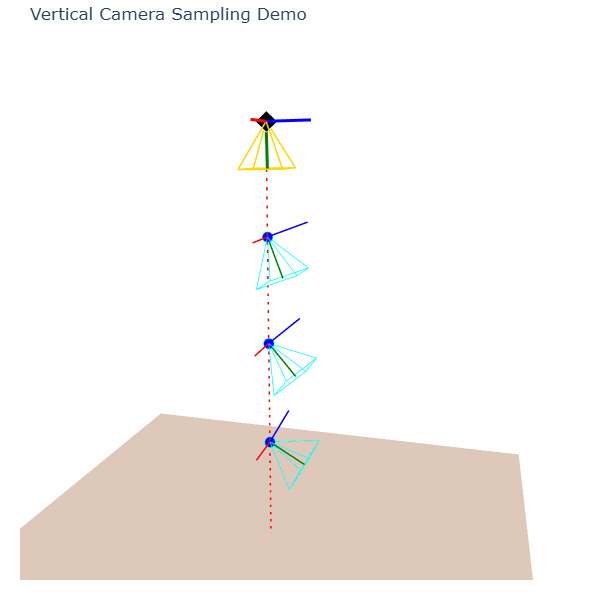}
    \includegraphics[width=0.24\linewidth]{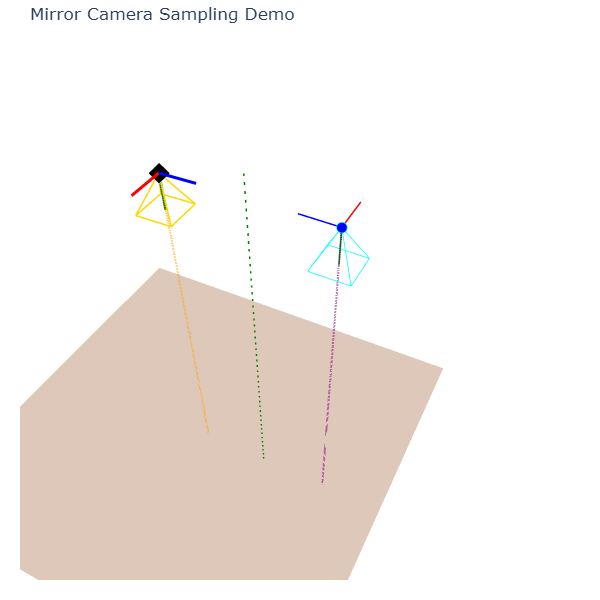}
    \includegraphics[width=0.24\linewidth]{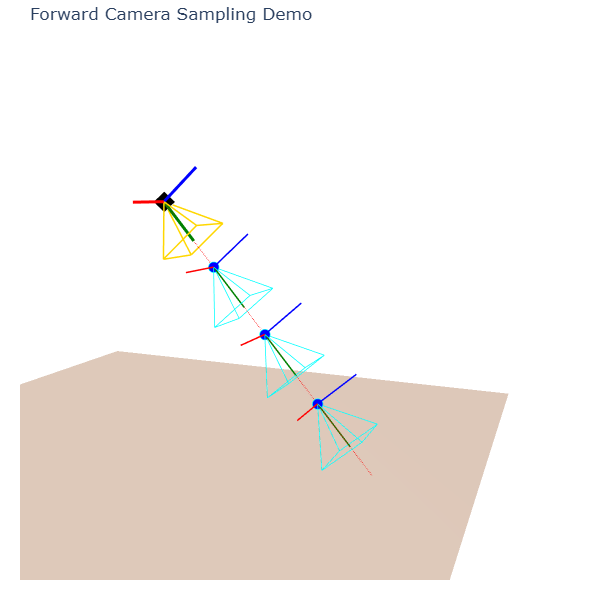}
    \includegraphics[width=0.24\linewidth]{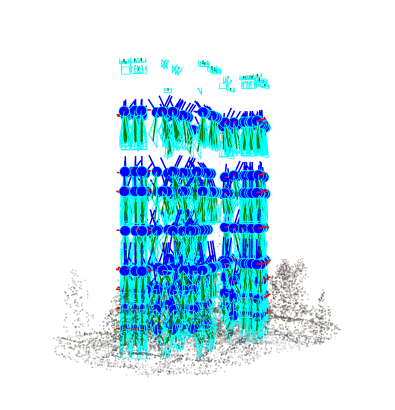}
    \caption{From left to right: Illustration of vertical, mirror, forward sampling, and visualization of complete synthetic camera generation for 20 randomly sampled real cameras in November 27, 2024 scene. Camera right, forward, and up directions are visualized in red, green, and blue, respectively. Ground plane is visualized in saddle brown. Dashed lines in vertical sampling and mirror sampling represent the vertical projection of the camera position to the ground plane. Dashed line in forward sampling represent the movement (forward) direction.}
    \label{fig:synthetic_camera_generation}
\end{figure}

While synthetic rendering solves temporal alignment, it inherits the viewpoint limitations of our real captures. Operational safety constraints prohibit UAV flights inside construction fence boundaries and below 65m crane height during weekdays, restricting our LiDAR missions to predominantly nadir perspectives captured from safe altitudes. Yet robust 2DCD algorithms require training and evaluation on diverse viewpoints - including ground-level views and oblique angles - to generalize across deployment scenarios. To transcend these physical limitations, we implement three complementary strategies addressing altitude restrictions, interior access constraints, and close-range imaging limitations:
(1) Vertical sampling: Generates cameras at varying altitudes below operational flight levels, perpendicular to the ground plane;
(2) Mirror sampling: Creates interior viewpoints by reflecting exterior camera positions across site boundaries; and
(3) Forward sampling: Produces close-range perspectives by translating cameras toward construction elements.
Examples are shown in Fig. \ref{fig:synthetic_camera_generation}.
This cascaded synthesis pipeline generates 32 synthetic viewpoints from each real camera position, effectively mitigating the viewpoint limitations imposed by operational safety constraints.

\subsubsection{Maximizing Pixel Coverage}
\label{subsubsec:pixel_coverage}

\begin{figure}
    \centering
    \input{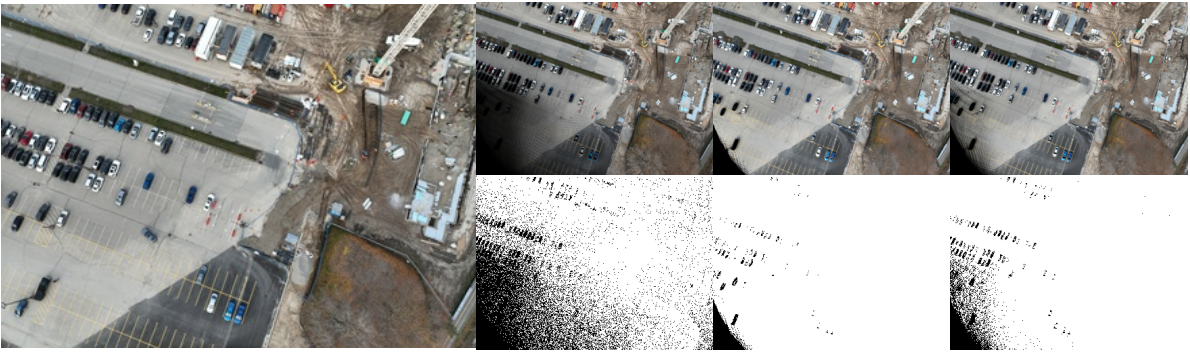}
    \caption{Left: original capture of the first image in the November 27 LiDAR mission. Right: synthetic rendering at the same scene, camera intrinsics and extrinsics, using (original resolution, 1 pixel point size), (1/8 resolution, 1 pixel point size), and (original resolution, 3 pixels point size), from left to right respectively. The respective pixel hit masks (white for pixel hit and black for pixel miss) are shown in the second row.}
    \label{fig:pixel_coverage}
\end{figure}

Despite our exceptionally dense LiDAR point clouds (200-800M points), their discrete, non-volumetric nature creates a fundamental rendering challenge - each point projects to isolated pixels rather than continuous surfaces. When projected to 2D at original resolution, we achieve only $\sim$77\% pixel coverage, leaving $\sim$23\% of pixels filled with zeros. These coverage gaps require explicit masking during neural network training and degrade synthetic image quality. Existing rendering solutions prove inadequate for our scale: open-source libraries either fail computationally on our massive point clouds \cite{open3d} or require mesh representations that introduce additional preprocessing overhead and potential reconstruction artifacts \cite{blender}.

We implement a pure PyTorch-based rendering pipeline that provides complete control over the synthesis process while fully exploiting our high-precision LiDAR data.
Our analysis quantifies pixel coverage improvements through two straightforward yet effective strategies. The first strategy to increase pixel coverage is to simply render at a lower resolution (downscale the camera intrinsics when projecting points to image plane). The second strategy is to make a simple model for volumetric points by letting each point occupy pixels in a disk of diameter 3 pixels.

Fig. \ref{fig:pixel_coverage} shows that although rendering at original resolution and 1 pixel point size (77.29\% pixel coverage) provides close-to-real texture on the top-right diagonal, the bottom-left diagonal has dark regions because of insufficient pixel coverage, and the missed pixels were assigned 0 values, which makes black colors. Rendering at 1/8 scale with 1 pixel point size (97.46\% pixel coverage) or rendering at original resolution and 3 pixels point size (94.05\% pixel coverage) both significant clears up the missing pixels due to point cloud \textit{sparsity}. However, regions of point cloud \textit{incompleteness} cannot be solved.
Our image resolution is at (5280, 3956) width and height in the raw data, (5645, 4082) in the undistorted images from COLMAP \cite{colmap}, and (706, 510) after 1/8 downscale, which is still higher than most existing 2DCD datasets \cite{whu_cd, dsamnet} and what most algorithms process \cite{fc_siam, change_former, change_mamba}.
In our iVISION-2DCD generation, we stick to using 1/8 resolution and 1 pixel point size.


\subsection{Semantic Segmentation and Change Labels}
\label{subsec:seg_and_change}

Generating accurate ground truth labels for construction 2DCD datasets faces three fundamental challenges: 2D images lack geometry information necessary for distinguishing structural changes from perspective variations~\cite{siam_kpconv, urb3dcd}, dense bi-temporal change annotation proves prohibitively expensive at scale~\cite{change_star_v1, i3pe}, and construction datasets require domain-specific filtering to exclude irrelevant environmental noise. To address these challenges, we perform semantic segmentation directly on 3D point clouds to preserve full geometric information, follow the single-temporal paradigm to annotate single-temporal semantic labels then deriving changes through bi-temporal comparison, and implement construction-focused annotation to filter irrelevant noise. This subsection details our semantic segmentation and change detection labeling methods.

\subsubsection{Semantic Segmentation Labels}

We use DJI Modify \cite{dji_modify} for initial automated point cloud labeling, followed by manual refinement to add construction-specific categories and correct labeling around dynamic objects. Additionally, we designate all regions outside the construction fence boundary as ``Ignore'' class, along with irrelevant temporal variations such as airborne snow, vegetation growth, and activities beyond site perimeters, ensuring our dataset focuses exclusively on construction-specific changes within the active construction zone. This approach achieves highly accurate per-point semantic labels with fine-grained class distinctions, leading to a dataset specifically tailored for construction automation research. Examples are shown in Fig. \ref{fig:first_figure}.

\subsubsection{Change Detection Labels}

Point clouds inherently produce asymmetric change labels due to varying point distributions \cite{urb3dcd,slpccd} - changes detected from time $t_1$ to $t_2$ differ from those detected from $t_2$ to $t_1$. Yet typical 2DCD algorithms typically require symmetric change masks where pixel $(i,j)$ shows identical change regardless of temporal direction \cite{whu_cd, levir_cd, dsamnet}. We resolve this fundamental mismatch through bi-directional 3D semantic comparisons, computing changes in both temporal directions, projecting and rasterizing each to image space, then taking union (logical OR) of the two resulting change maps as the final 2DCD ground-truth label. This ensures the resulting 2D  maintain the symmetry that canonical 2DCD training demands.

\section{Dataset Characteristics}
\label{sec:characteristics}

We present the visual quality challenge descriptions and examples and dataset statistics in this section.

\subsection{Challenging Cases}

Our dataset exhibits 4 challenging cases: Snow, Ignore Ground, Dynamic Objects, and RGB Projection Errors.

\begin{figure}
    \centering
    \includegraphics[width=\textwidth]{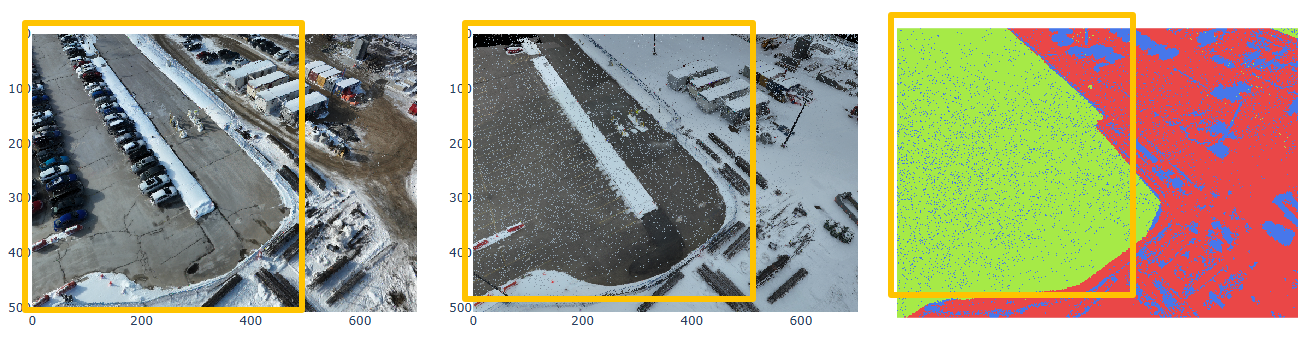}
    \caption{Example of airborne snow captured in the point clouds. Left column: time $t_{1}$ input images. Middle column: time $t_{2}$ input images. Right column: change labels. Changes due to airborne snow. This has been re-mapped to ``Ignore'' class and not present in our generated ground-truth changes.}
    \label{fig:snow}
\end{figure}

\textbf{Snow} creates annotation ambiguity as accumulated snow intermixes with construction elements while airborne snow remains separable. We assign accumulated snow the semantic class of underlying objects and label airborne particles as ``Snow''. Examples are shown in Fig. \ref{fig:snow}.

\begin{figure}
    \centering
    \includegraphics[width=0.72\textwidth]{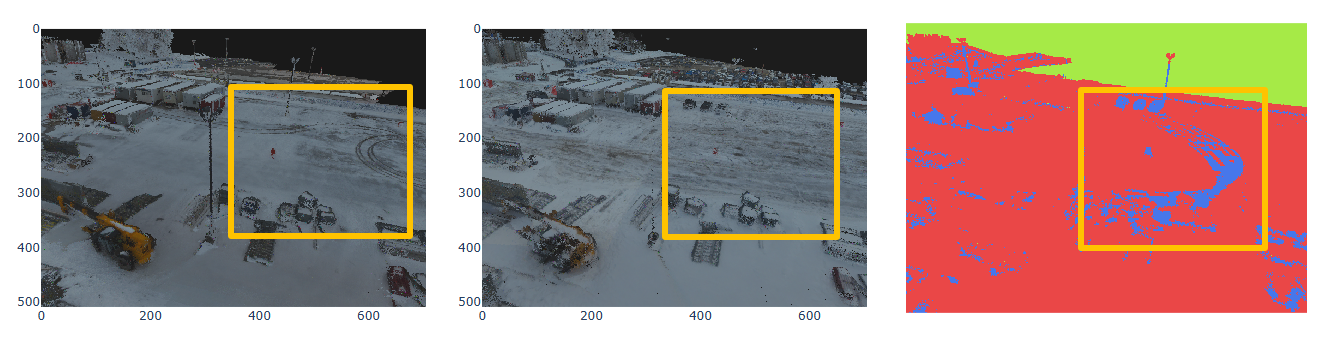}
    \includegraphics[width=0.24\textwidth]{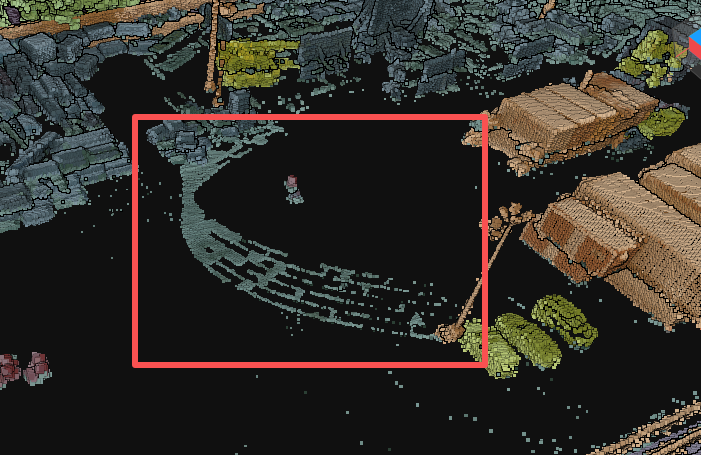}
    \caption{Example of ignore ground semantic changes hard to recognize from RGB inputs. From left to right: time $t_{1}$ input image, time $t_{2}$ input image, ground truth change map, and semantic segmentation map (with ``Ground'' class made invisible).}
    \label{fig:ignore_ground}
\end{figure}

\textbf{Ignore Ground.} Subtle terrain variations from material staging and excavation are invisible in RGB but indicate construction progress through micro-topographical changes. Examples are shown in Fig. \ref{fig:ignore_ground}.

\begin{figure}
    \centering
    \includegraphics[width=0.24\textwidth]{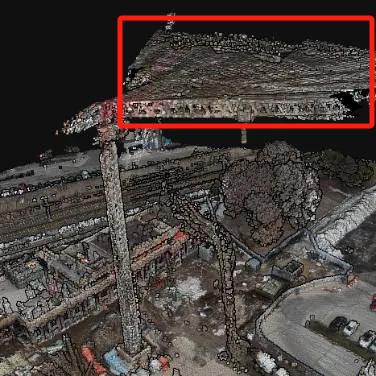}
    \includegraphics[width=0.24\textwidth]{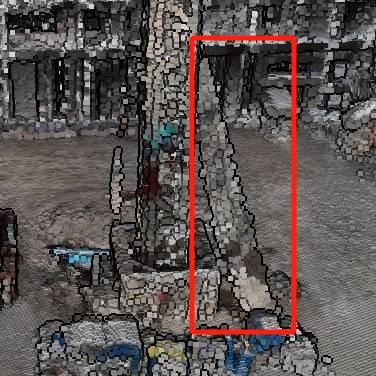}
    \includegraphics[width=0.24\textwidth]{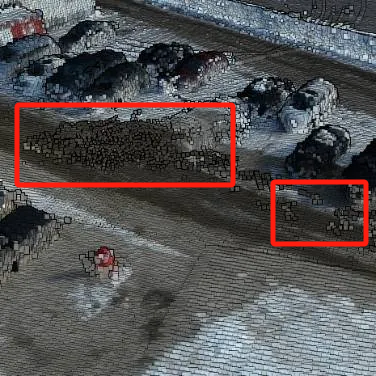}
    \includegraphics[width=0.24\textwidth]{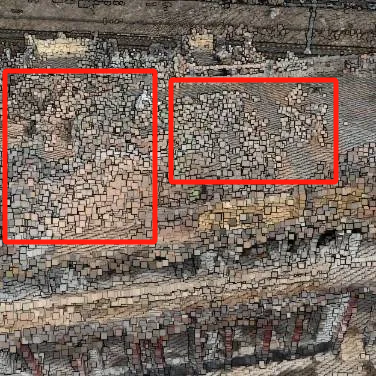}
    \caption{Examples of dynamic elements in the construction site. Figures shown by DJI Modify to amplify the dynamic points. From left to right: moving crane, moving string attached to the crane, moving vehicle, and moving construction workers.}
    \label{fig:dynamic_objects}
\end{figure}

\textbf{Dynamic Objects} create point cloud artifacts: noise clusters from slow objects (cranes, workers), trailing artifacts from vehicles, and missing data from high-speed traffic ($\geq$ 60 km/h). Examples are shown in Fig. \ref{fig:dynamic_objects}.

\begin{figure}
    \centering
    \includegraphics[width=0.32\textwidth]{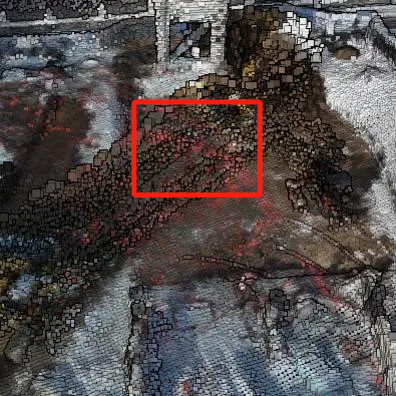}
    \includegraphics[width=0.32\textwidth]{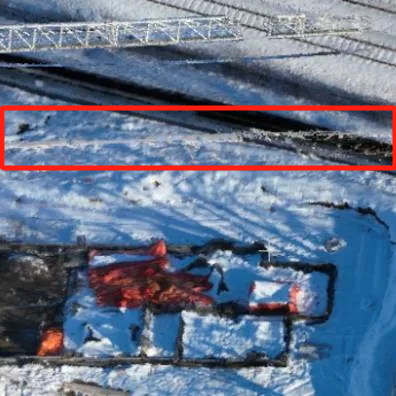}
    \includegraphics[width=0.32\textwidth]{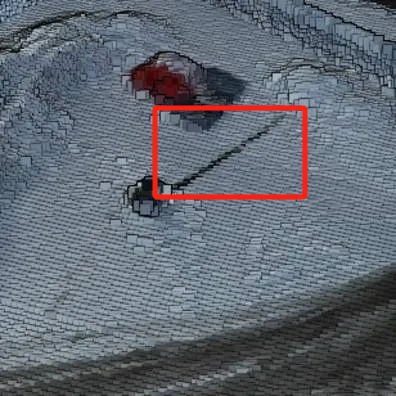}
    \caption{Examples of RGB projection error in the reconstructed LiDAR point clouds. Figures shown by DJI Modify to amplify the error points. From left to right: red color of the crane falsely projected onto the ground; crane frameworks grey colors projected onto the ground; light pole in construction site texture projected onto the ground.}
    \label{fig:rgb_projection}
\end{figure}

\textbf{RGB Projection Errors.} The LiDAR reconstruction pipeline introduces systematic complexities where thin elevated structures (cranes, poles) project incorrect colors onto ground surfaces, creating confounding visual patterns. Examples are shown in Fig. \ref{fig:rgb_projection}.

\subsection{Dataset Statistics}

For the temporal window spanning November 27, 2024 to March 12, 2025 (17 scenes), we generate 16 consecutive scene pairs, yielding 32 bi-directional change maps. Our drone LiDAR missions typically contains 300-350 images per scene. Therefore, we downsample the $32N$ synthetic cameras generated from $N$ original cameras to 300 per scene, resulting in $300\times16=4800$ synthetic cameras.

\section{Experiments}

\begin{table}[]
    \centering
    \caption{Benchmark results of widely-studied and state-of-the-art methods on our iVISION-2DCD dataset and five other well-established benchmark datasets. All scores are mean Intersection over Union (mIoU) in percentages. Numbers in parentheses indicate rankings on each dataset. Top three methods on each dataset are highlighted in bold. Methods are sorted by their performance on iVISION-2DCD in descending order.}
    \resizebox{\textwidth}{!}{%
    \begin{tabular}{lcccccc}
\hline
Method & LEVIR-CD \cite{levir_cd} & SYSU-CD \cite{dsamnet} & OSCD \cite{oscd} & CDD \cite{cdd} & AirChange \cite{airchange} & iVISION-2DCD (Ours) \\
\hline
FTN & 33.05 (16) & \textbf{72.24 (2)} & 25.73 (16) & 86.18 (4) & 33.53 (18) & \textbf{53.77 (1)} \\
ChangeMamba & \textbf{84.67 (1)} & \textbf{72.26 (1)} & \textbf{48.97 (2)} & \textbf{91.35 (1)} & \textbf{73.87 (3)} & \textbf{52.56 (2)} \\
HCGMNet & 68.44 (5) & 70.58 (4) & 47.93 (13) & \textbf{88.72 (3)} & 71.98 (4) & \textbf{52.05 (3)} \\
BiFA & \textbf{81.11 (2)} & 61.03 (10) & 48.62 (8) & 44.31 (14) & \textbf{75.79 (1)} & 50.55 (4) \\
DsferNet & 32.00 (17) & \textbf{72.03 (3)} & 48.70 (5) & \textbf{91.14 (2)} & 50.81 (15) & 46.56 (5) \\
RFL-CDNet & 62.47 (6) & 65.64 (8) & 48.63 (7) & 50.77 (9) & 63.84 (10) & 44.32 (6) \\
CDXFormer & \textbf{76.55 (3)} & 68.05 (6) & 48.62 (9) & 84.56 (5) & 68.39 (6) & 43.49 (7) \\
ChangeFormer & 54.58 (8) & 63.21 (9) & \textbf{50.38 (1)} & 44.91 (13) & \textbf{75.07 (2)} & 41.79 (8) \\
SNUNet-ECAM & 48.75 (11) & 68.69 (5) & \textbf{48.70 (3)} & 57.31 (8) & 70.24 (5) & 41.71 (9) \\
FC-Siam-diff & 6.82 (18) & 55.43 (12) & 2.93 (18) & 43.99 (15) & 61.81 (12) & 40.98 (10) \\
ChangeNext & 49.16 (9) & 52.70 (16) & 46.66 (14) & 61.72 (7) & 57.20 (13) & 40.36 (11) \\
FC-Siam-conc & 48.23 (12) & 54.01 (14) & 48.70 (4) & 43.81 (16) & 64.54 (9) & 40.10 (12) \\
HANet & 68.65 (4) & 66.18 (7) & 48.11 (12) & 45.27 (11) & 68.11 (7) & 39.82 (13) \\
TinyCD & 43.44 (14) & 54.85 (13) & 22.34 (17) & 65.82 (6) & 54.44 (14) & 39.41 (14) \\
FC-EF & 40.38 (15) & 53.36 (15) & 48.62 (11) & 43.23 (17) & 67.27 (8) & 39.16 (15) \\
DSIFN & 48.99 (10) & 40.12 (17) & 48.62 (10) & 44.92 (12) & 49.81 (16) & 38.73 (16) \\
DSAMNet & 47.46 (13) & 31.79 (18) & 48.70 (6) & 39.64 (18) & 41.75 (17) & 35.73 (17) \\
Changer & 54.88 (7) & 59.30 (11) & 40.14 (15) & 48.60 (10) & 61.90 (11) & 29.78 (18) \\
\hline
\end{tabular}

    }
    \label{tab:benchmarks}
\end{table}

\subsection{Setup}

\textbf{Selected Methods and Datasets. } We evaluate 18 representative 2D change detection methods: FC-Siam-diff, FC-Siam-conc, FC-EF~\cite{fc_siam}, DSAMNet~\cite{dsamnet}, DSIFN~\cite{dsifn}, HANet~\cite{hanet}, SNUNet-ECAM~\cite{snunet_cd}, TinyCD~\cite{tinycd}, RFL-CDNet~\cite{rfl_cdnet}, ChangeFormer~\cite{change_former}, BiFA~\cite{bifa}, CDXFormer~\cite{cdxformer}, Changer~\cite{changer}, FTN~\cite{ftn}, HCGMNet~\cite{hcgmnet}, ChangeMamba~\cite{change_mamba}, ChangeNext, DsferNet~\cite{dsfer_net}. We benchmark these methods on six datasets: our oblique-view construction dataset iVISION-2DCD alongside five established nadir-view benchmarks - LEVIR-CD~\cite{levir_cd}, SYSU-CD~\cite{dsamnet}, OSCD~\cite{oscd}, CDD~\cite{cdd}, and AirChange~\cite{airchange}. These methods and datasets were selected based on their popularity, reported performance, and source code availability.

\textbf{Training. } We set up our training following \cite{change_star_v1,fc_siam,i3pe}. We assign class weights inverse-proportionally to the number of pixels in the dataset. Throughout all experiments, we use SGD optimizer with initial learning rate $10^{-3}$, momentum $0.9$, and weight decay $10^{-4}$. We use polynomial learning rate scheduler with power $0.9$. Batch size was set to 4 throughout. Standard data augmentations including random rotation, random flip, random color jitter, and random crop has been applied on all existing datasets. Differently, iVISION-2DCD synthetically samples diverse close-up views towards the construction site. Random rotation is no longer a reasonable data augmentations as it would create clearly out-of-distribution samples, e.g., an upside-down crane, and hence removed from the list.

\subsection{Results}

Table \ref{tab:benchmarks} reports mIoU scores across all methods and datasets, revealing that iVISION-2DCD presents fundamentally different challenges compared to existing benchmarks.

\textbf{Performance Degradation.} All methods experience significant performance drops on iVISION-2DCD compared to established benchmarks. State-of-the-art methods like ChangeMamba decline from 84.67\% (LEVIR-CD) and 91.35\% (CDD) to 52.56\% on our dataset, while BiFA drops from 81.11\% (LEVIR-CD) to 50.55\%.

\textbf{Ranking Inversions.} More critically, method rankings exhibit dramatic reversals between datasets. Methods achieving top performance on existing benchmarks often fail on iVISION-2DCD: ChangeFormer ranks 1st on OSCD (50.38\%) and 2nd on AirChange (75.07\%) but drops to 8th place (41.79\%) on ours. DsferNet ranks 2nd on CDD (91.14\%) and 3rd on SYSU-CD (72.03\%) yet only 5th on ours (46.56\%) while failing catastrophically on LEVIR-CD (17th, 32.00\%). Conversely, methods performing poorly elsewhere succeed on our dataset: FTN consistently ranks near bottom positions (18th on AirChange, 16th on LEVIR-CD and OSCD) yet achieves top performance (53.77\%) on iVISION-2DCD, while FC-Siam-diff catastrophically fails on LEVIR-CD (18th, 6.82\%) and OSCD (18th, 2.93\%) but manages 10th place (40.98\%) on ours.

These dramatic ranking inversions indicate that iVISION-2DCD captures unique visual complexities from oblique-view construction scenes that fundamentally differ from nadir-view patterns in existing aerial and satellite benchmarks.

\section{Conclusion}

In this paper, we have presented iVISION-2DCD, a comprehensive synthetic 2DCD dataset for construction monitoring generated from dense LiDAR point clouds. We developed novel view synthesis techniques to overcome fundamental challenges in 2DCD data curation: achieving pixel-perfect bi-temporal alignment through synthetic rendering, transcending operational viewpoint limitations via systematic camera generation, and optimizing pixel coverage to 97\% through strategic resolution choices. Our semi-automated semantic segmentation pipeline leverages single-temporal labeling paradigms to generate accurate change maps while preserving challenging real-world cases including snow covering, terrain changes invisible in RGB, dynamic object artifacts, and systematic projection errors. The dataset spans over one year with 50+ temporal captures, providing both geometric and textural richness essential for robust algorithm development. Through extensive experiments with widely-studied and state-of-the-art 2DCD algorithms, we demonstrated that iVISION-2DCD introduces novel challenges specific to construction monitoring applications. Our work establishes a new benchmark for viewpoint-robust 2DCD in construction environments, opening advancement opportunities at the intersection of computer vision, robotics, and construction automation.



\section*{Acknowledgment}

This work was supported by National Research Council of Canada (NRC) through
the Construction Sector Digitalization and Productivity Challenge program, project number CSDP-014-1.


\end{document}